\documentclass[journal]{IEEEtran}

\usepackage{array}

\usepackage{float}
\usepackage{makecell}
\usepackage{graphicx}
\usepackage{multirow}
\usepackage{cite}
\usepackage{amsmath,amssymb,amsfonts}
\usepackage{algorithmic}
\usepackage{graphicx}
\usepackage{textcomp}
\usepackage{xcolor}
\usepackage[utf8]{inputenc}
\usepackage{amsmath, amssymb}
\usepackage{enumitem}
\usepackage{xcolor}
\usepackage{lipsum}
\usepackage{comment}

\usepackage{placeins}

\begin{document}

\title{Evaluation of a Multi-Link Machine Learning Approach for Rainfall Detection Using Satellite Microwave Links}

\author{
    \IEEEauthorblockN{
    Mostafa Ftouni\textsuperscript{1},
     Margherita Spalla\textsuperscript{2},
     Matteo Colli\textsuperscript{2},
     Andrea Randazzo\textsuperscript{1},
     Christian Gianoglio\textsuperscript{1}
    }\\
    \IEEEauthorblockA{\textsuperscript{1}DITEN, University of Genoa, Italy.\\
    \textsuperscript{2}Artys, Darts Engineering, Genoa, Italy.
    }
}

\maketitle

\begin{abstract}
The impacts of climate change have increased the need for reliable real-time rainfall monitoring systems. This study investigates rainfall detection using a network of opportunistic sensors based on satellite microwave links (SMLs) combined with machine learning (ML) classifiers. Two models, a random forest (RF) and a convolutional neural network (CNN), were trained on aggregated SML data, including information from neighboring links. The results demonstrate that leveraging information from multiple links substantially enhances the detection of both wet and dry periods compared to single-link approaches. Moreover, the RF model consistently provides a more stable and reliable trade-off than CNN, improving rainfall detection and reducing misclassifications overall, with particularly strong performance during harsh weather conditions. The study spanned up to 32 rainy days between 2021 and 2024.

\end{abstract}

\begin{IEEEkeywords}
satellite microwave link, machine learning, rainfall monitoring, neighboring links.
\end{IEEEkeywords}

\IEEEpeerreviewmaketitle


\section{Introduction}
\label{Introduction}

\IEEEPARstart{A}{ccurate} real-time rainfall monitoring is essential for water resource management, agriculture, urban planning, and disaster prevention, supporting early warning systems for hazards such as floods and landslides, reducing human and economic losses \cite{mirus2018integrating}, \cite{FORTELLI2019}, \cite{ZHAO2022106456}. However, achieving sufficient spatial and temporal coverage remains challenging. Opportunistic microwave links provide a cost-effective solution with wide spatial coverage via existing telecommunication infrastructures \cite{chwala2012precipitation}, \cite{chwala2019commercial}, \cite{giannetti2021opportunistic}, \cite{giannetti2023opportunistic}. Satellite microwave links (SMLs), especially in the Ku and Ka bands, are sensitive to rainfall-induced attenuation, enabling indirect rainfall estimation \cite{mercier2015estimation}, \cite{giannetti2017real}. An example is the Smart Rainfall System (SRS) \cite{colli2018field}, deployed in the Polcevera River Basin, which uses satellite signal attenuation to classify wet/dry periods and estimate rainfall intensity (RI).

In recent years, ML and signal processing techniques have been widely explored for the real-time recognition of rain and non-rain conditions. In \cite{giro2022real}, a randomized trees classifier was applied to received power signals for precipitation detection and attenuation estimation, enabling rain rate retrieval at 1-minute resolution. In \cite{xian2020rainfall}, a support vector machine (SVM) was used to classify rain and non-rain periods from SML signals, while a long short-term memory network estimated the attenuation baseline. A convolutional neural network (CNN) was proposed in \cite{polz2020rain} to detect rainfall events using data from 800 commercial microwave links (CML) in Germany, with the gauge-adjusted RADOLAN-RW radar product as reference. Similarly, \cite{oydvin2024simple} trained two neural networks based on signal attenuation measurements from CML, using rain-gauge and radar data as reference sources, to identify rainy and dry time steps across a large CML dataset. In \cite{song2020experimental}, an SVM was employed to classify wet and dry periods using attenuation statistics. However, despite the growing interest in SML-based rainfall monitoring, a significant gap remains in reconstructing rainfall fields over large areas using SML networks combined with ML techniques.

In our previous study \cite{gianoglio2024rainfall}, wet and dry periods were classified independently for each SRS link, using only features derived from the same link, thus neglecting the spatial distribution of sensors over the territory. This single-link approach, although effective, does not account for the spatial correlation of rainfall events, which typically affect multiple nearby links simultaneously.
In this paper, we address this limitation by introducing a spatially aware multi-link approach, which represents the main novelty of this work. The proposed method includes information from neighboring links, which are likely to be affected by the same rainfall event due to their geographical proximity, thus providing complementary information that enhances the robustness and accuracy of rain detection. This transition from an independent single-link analysis to a multi-link modeling strategy allows the model to better capture the spatial structure of rainfall phenomena, improving the classification of wet and dry periods compared to our previous work \cite{gianoglio2024rainfall}. The study was validated by using four groups of neighboring links of the Valpolcevera river basin, including a total of 16 SRS sensors, to evaluate both a feature-based RF model and a 2D-CNN fed by the raw signals.


\section{Data Acquisition And Methodology}
\label{Data Acquisition And Methodology}

\subsection{Dataset}
\begin{figure}[b]
\centerline{\includegraphics[width = 1\linewidth]{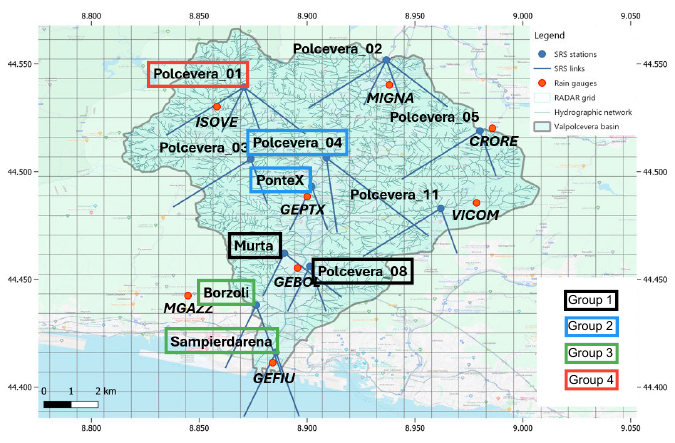}}
\caption{SRS links in Genoa utilized for data acquisition.}
\label{Map}
\end{figure}

The dataset used in this study was collected between January 2021 and August 2024 from 16 SRS sensors deployed across 7 sites (Fig. \ref{Map}), receiving microwave signal power from Ku-band satellite links. The measurements are affected by rainfall-induced attenuation along the propagation path. Further details on the SRS configuration and operation are provided in \cite{colli2018field}.
For this analysis, four groups of neighboring SRS links were defined, each comprising four spatially close links within the study area (Fig. \ref{Map}). The grouping was based on the spatial proximity of the available SRS receiving stations, where links were selected if they shared the same site or were located within approximately 3 km of each other. Links pointing toward different satellites are also included to exploit the joint temporal behavior of received-power variations across links. This empirical choice aims to group receiving stations that are sufficiently close to be affected by precipitation in the same local basin.

Table \ref{grouped_sites} summarizes the selected groups, including installation site and satellite orientation for each link.

For each group, rainfall-significant days were selected based on data availability across all four links, retaining only those with complete measurements. Days with missing data in any link were excluded to ensure synchronized multi-input observations and avoid inconsistencies. As a result, Groups 1, 2, 3, and 4 include 32, 30, 24, and 20 days, respectively. RI was derived from Mt. Settepani weather radar maps, located about 60 km from the study area \cite{gianoglio2024rainfall,colli2020rainfall}, with 5-min temporal resolution (288 maps/day). For each SRS link and day, RI values were extracted at the intersections between the link path and the radar grid, following \cite{gianoglio2024rainfall}. Wet/dry labels were then assigned by converting RI into binary classes (0: dry, 1: wet) according to the following rule: 
\begin{equation}
\centering
y =
\begin{cases}
0 & \text{if } RI < th_{RI} \\
1 & \text{if } RI \geq th_{RI}
\end{cases}
\label{first-equation}
\end{equation}
where $th_{RI}$ = 0.1 mm/h as in \cite{gianoglio2024rainfall}, \cite{gianoglio2023rain}.

As a result, the dataset utilized in this study can be formalized as:
\begin{equation}
\begin{aligned}
\mathcal{D}_{\text{link}} = &
\Big\{ (\boldsymbol{x}, \boldsymbol{y})_d \,\Big|\,
 \boldsymbol{x}_d \in \mathbb{R}^{288 \times 1},\ 
  \boldsymbol{y}_d \in \{0,1\}^{288 \times 1},\\[-4pt]
& d = 1,\ldots,N_{\text{d,link}} \Big\}
\end{aligned}
\label{second-equation}
\end{equation}
where the link denotes an SRS link, and $N_{d,\text{link}} \in {32, 30, 24, 20}$ represents the number of available days, as reported in Table \ref{grouped_sites}.

\begin{table}[b]
\caption{Neighboring SRS Link Groups and their Available Days.}
\centering
\setlength{\tabcolsep}{3pt}
\renewcommand{\arraystretch}{0.1} 
\begin{tabular}{|c|c|c|c|}
\hline
\textbf{Group ID} & \textbf{Number of Days} & \textbf{Site} & \textbf{Satellites} \\
\hline
\multirow{2}{*}{\textbf{1}} 
& \multirow{2}{*}{32} 
& \makecell[c]{Murta \\ (MT, ME)}
& \makecell[c]{Turksat (42E) \\ Eutelsat (5W)} \\ \cline{3-4}
& & \makecell[c]{Polcevera-08 \\ (P8A, P8E)}
& \makecell[c]{Astra (19.2E) \\ Eutelsat (5W)} \\
\hline
\multirow{2}{*}{\textbf{2}} 
& \multirow{2}{*}{30} 
& \makecell[c]{Polcevera-04 \\ (P4T, P4H)}
& \makecell[c]{Turksat (42E) \\ Hot Bird (13E)} \\ \cline{3-4}
& & \makecell[c]{PonteX \\ (PXA, PXE)}
& \makecell[c]{Astra (19.2E) \\ Eutelsat (5W)} \\
\hline
\multirow{2}{*}{\textbf{3}}
& \multirow{2}{*}{24} 
& \makecell[c]{Sampierdarena \\ (SE, SA)}
& \makecell[c]{Eutelsat (5W) \\ Astra (19.2E)} \\ \cline{3-4}
& & \makecell[c]{Borzoli \\ (BA, BE)}
& \makecell[c]{Astra (19.2E) \\ Eutelsat (5W)} \\
\hline
\multirow{2}{*}{\textbf{4}} 
& \multirow{2}{*}{20} 
& \makecell[c]{Polcevera-01-East \\ (P1T, P1A)}
& \makecell[c]{Turksat (42E) \\ Astra (19.2E)} \\ \cline{3-4}
& & \makecell[c]{Polcevera-01-West \\ (P1E, P1H)}
& \makecell[c]{Eutelsat (5W) \\ Hispasat (30W)} \\
\hline
\end{tabular}
\label{grouped_sites}
\end{table}
\subsection{Classification Algorithms}

Two ML models were considered in this study, an RF model and a 2D-CNN. The choice of the 2D-CNN was motivated by the multi-link data, enabling the capture of both temporal and spatial dependencies, unlike 1D-CNN which models only temporal information. Both models were evaluated using the multi-input (MI) approach, where data from the target link and its neighboring links in the same group are jointly used to predict the target wet/dry state. Each link is alternately considered as the target, and a separate model is trained and tested for each case. For comparison, RF was also evaluated using the single-input (SI) approach, which follows \cite{gianoglio2024rainfall} and relies only on the target-link data.

A moving window of length $l \in \{15, 30\}$ minutes was applied to the SRS signal $\boldsymbol{x}_d \in \mathcal{D}_{\text{link}}$. For each observation $x_i$, a time-series array $\boldsymbol{w}_i = [x_{i-l+1}, \ldots, x_i]$ was constructed including the previous $l-1$ observations together with the current  Radar-derived RI, available at 5-minute resolution, was used as ground truth, while SRS measurements were sampled at 1-minute resolution. Samples were defined at radar timestamps, where the corresponding RI value provided the label, and each input window was formed from the preceding $l$ minutes of SML data, ensuring temporal alignment and consistency.
For the 2D-CNN model, the raw time-series array $\boldsymbol{w}_i$ is used directly as the model input. In contrast, the RF classifier uses a vector of 11 features extracted from $\boldsymbol{w}_i$, following the previous study \cite{gianoglio2024rainfall}. In the MI approach, feature vectors from neighboring links were concatenated and processed by the RF model, while the 2D-CNN directly exploited the time-series arrays from the different links as a two-dimensional input.

The ML training dataset is formalized as:
\begin{equation}
\begin{aligned}
\mathcal{T}^{RF}_{\text{link}} = &
\Big\{ (W, \boldsymbol{y})_d \,\Big|\,
 W_d \in \mathbb{R}^{288 \times (N_f * N_l)} \\ 
&  \boldsymbol{y}_d \in \{0,1\}^{288 \times 1}, d = 1,\ldots,N_{\text{d,link}} \Big\}
\end{aligned}
\label{third-equation}
\end{equation}
and
\begin{equation}
\begin{aligned}
\mathcal{T}^{2D\text{-CNN}}_{\text{link}} = &
\Big\{ (\mathcal{W}, \boldsymbol{y})_d \,\Big|\,
 \mathcal{W}_d \in \mathbb{R}^{288 \times l \times N_l},\\
&  \boldsymbol{y}_d \in \{0,1\}^{288 \times 1},
d = 1,\ldots,N_{\text{d,link}} \Big\}
\end{aligned}
\label{Fourth-equation}
\end{equation}

\noindent where $N_f = 11$ represents the number of features, and $N_l = 4$ represents the number of links.

\subsection{Training Procedure}

A leave-one-out cross-validation technique was employed to prevent model overfitting mainly arising from the autocorrelation of rainfall events and to ensure evaluation on entirely unseen daily sequences. For each link, one day from $\mathcal{T}^{RF}_{\text{link}}$ and $\mathcal{T}^{2D\text{-CNN}}_{\text{link}}$ was iteratively used for testing, while the remaining days were used for training, ensuring more reliable generalization assessment. In this study, the implementation relies on complete data, whereas operational use may require handling missing inputs. The training set was further split into training and validation subsets, comprising 80\% and 20\% of the data, respectively.
Since non-rainy observations were predominant, class weights were computed prior to model training and consistently applied to both the RF and 2D-CNN models to address class imbalance and ensure a fair comparison. In this approach, a higher weight was assigned to the minority class, so that misclassification errors on rainy samples were penalized more heavily.

The validation set was used to tune the hyperparameters of RF and 2D-CNN. For the RF model, the hyperparameters considered were the number of trees 
$N_{\text{tree}} = \{50,\,100,\,200\}$, the maximum tree depth 
$\text{max\_depth} = \{10,\,20,\,\text{None}\}$, the minimum number of samples required to perform a split $\text{min\_sam} = \{2,\,5\}$, and the minimum number of samples required to form a leaf node $\text{min\_leaf} = \{1,\,5\}$.
For the 2D-CNN model, the hyperparameter search explored a compact set of architectures, characterized by a limited number of convolutional blocks, which is well suited to the MI approach given the short temporal window ($w = 15$ or $30$) and the limited size of the dataset.
The tested configurations included the number of convolutional filters in the first layer ${4, 8, 16}$, with the number of filters doubling in each subsequent convolutional block, the number of convolutional blocks ${2, 3}$, a kernel size of $(3, 2)$, a fixed batch size of 64, and a learning rate of 0.001. Each architecture followed a standard Conv2D--BatchNorm--Pooling design, ending with global average pooling and two dense layers: a 64-neuron hidden layer and an output layer for classification.
Hyperparameter tuning was performed by evaluating each configuration across all days, computing the validation Harmonic Mean (HM) of recall and specificity, and selecting the configuration that achieved the highest average HM. The selected configuration was then fixed for testing. For both the RF and 2D-CNN models, the predicted output represents the probability of rainfall occurrence. Therefore, the validation set was used to evaluate multiple decision thresholds and select the value that provides the best wet/dry discrimination, as described in the next section.

\section{Results}
\label{Results}

Three performance metrics were adopted following \cite{gianoglio2024rainfall}. Specificity ($\text{Spe}=\frac{\text{TN}}{\text{TN}+\text{FP}}$), Recall ($\text{Rec}=\frac{\text{TP}}{\text{TP}+\text{FN}}$), and their harmonic mean ($\text{HM}=\frac{2\text{Spe}\text{Rec}}{\text{Spe}+\text{Rec}}$), were used to ensure robustness to class imbalance. Here, TP/TN denote correct wet/dry classifications, while FP/FN indicate misclassifications. All metrics range from 0 to 1.

\begin{figure}[b]
\centerline{\includegraphics[width = 1\linewidth]{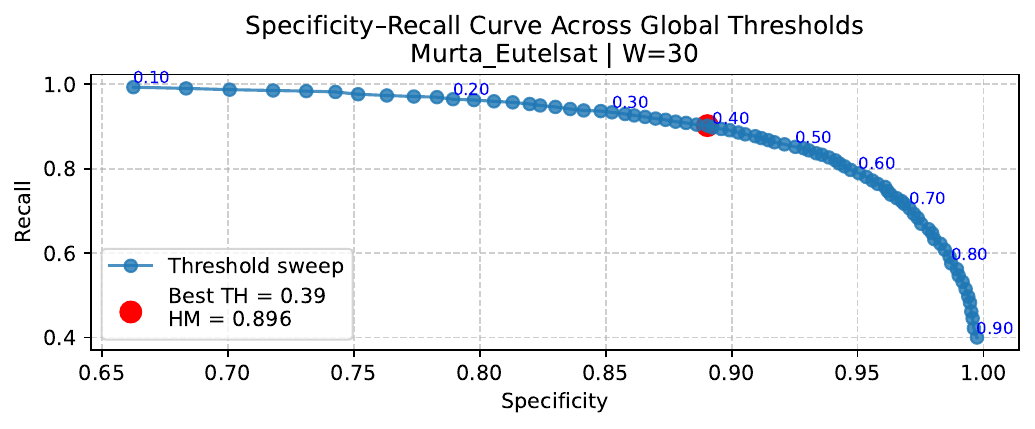}}
\caption{Recall–specificity curve for RF on the Murta–Eutelsat link using a 30-minute window, evaluated over thresholds ranging from 0.1 to 0.9.}
\label{figure-curve}
\end{figure}

Figure \ref{figure-curve} presents an example of the Rec–Spe curve over thresholds ranging from 0.1 to 0.9, highlighting that the threshold yielding the highest HM corresponds to 0.39. The optimal threshold is selected using a validation set aggregated across all evaluation days, after removing duplicate samples to ensure independence. The threshold maximizing validation HM is selected and fixed for testing.

\begin{figure}[tb]
\centerline{\includegraphics[width = 1\linewidth]{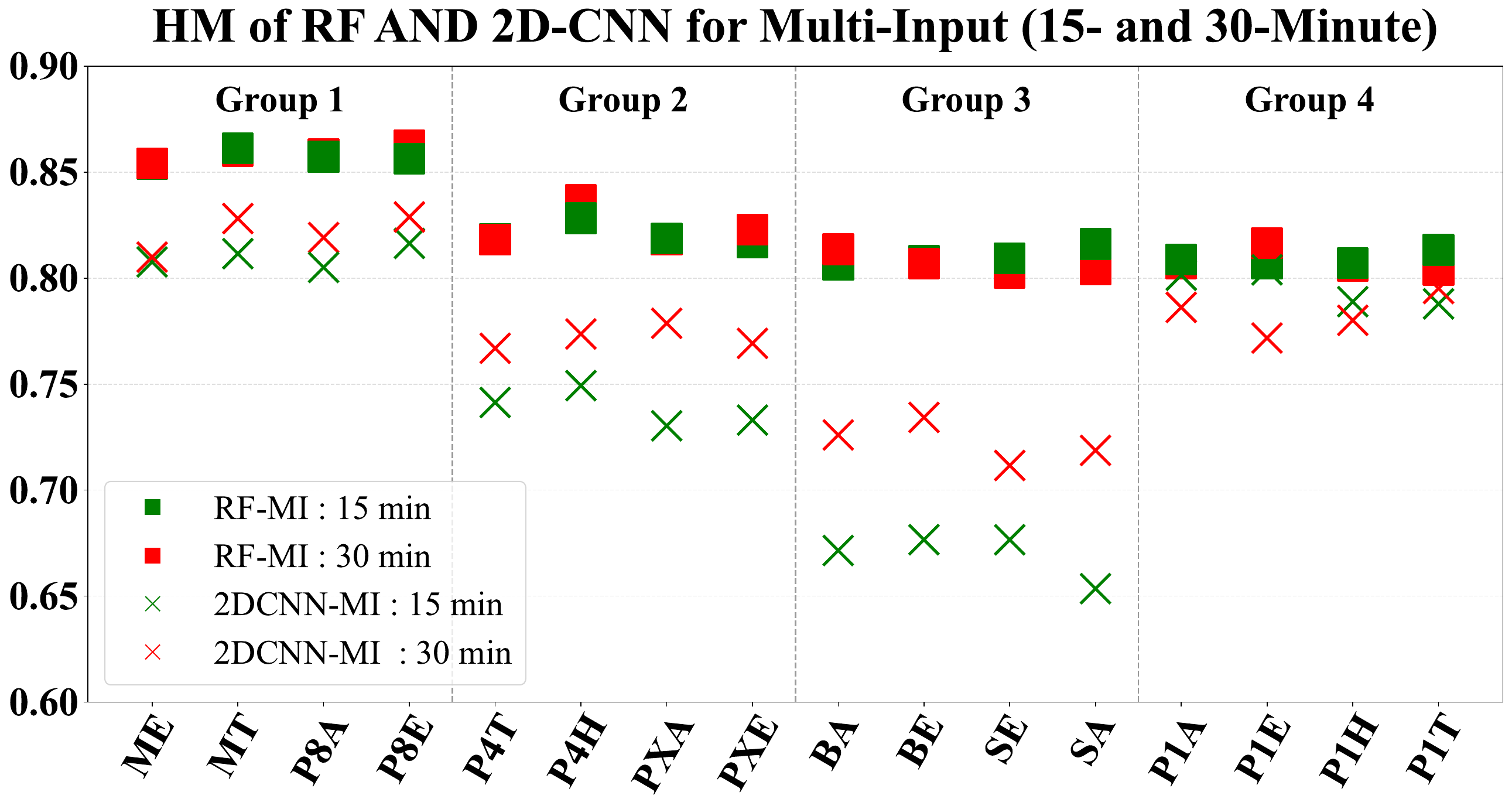}}
\caption{Harmonic mean of rain detection for RF and 2D-CNN using the multi-input approach, evaluated over 15- and 30-minute time windows. The names on the x axis refer to the single link as listed in Table \ref{grouped_sites}.}
\label{figure-cnn}
\end{figure}

Figure \ref{figure-cnn} illustrates the performance of the RF and 2D-CNN models under the MI input approach. Both models were evaluated using two temporal windows, 15 and 30 minutes, allowing to assess the influence of spatial information and window length on classification performance. The results show that the RF model performs largely similarly when using either window across all links. In contrast, the 2D-CNN generally benefits from the larger 30-minute window, particularly for Groups 2 and 3, where the difference between the two windows ranges from 3 \% to 7 \%. For the remaining groups, the performance gap is less pronounced. Despite this improvement, the RF model continues to provide a more balanced prediction and demonstrates greater stability across links. An exception is observed in Group 4, where the 2D-CNN performance approaches that of RF; however, RF still achieves slightly better overall results.

\begin{table}[tb]
\centering
\caption{RF and 2D-CNN MI Performance for Group 1 Across 15- and 30-Minute Windows (Rec at Three Rain Levels, Spe, and HM).}
\setlength{\tabcolsep}{3pt}
\renewcommand{\arraystretch}{0.9}
\begin{tabular}{c c c c c c c c}
\hline
Link & Model & Window & Rec-L & Rec-M & Rec-H & Spe & HM \\
\hline
MT & RF-MI & 15 & 80 & 98 & 99.5 & 86 & 86.1 \\
   &    & 30 & 79.4 & 97.2 & 99 & 86.4 & 85.9 \\
   & 2D-CNN-MI & 15 & 73.4 & 89.3 & 98.5 & 81 & 81.2 \\
   &    & 30 & 78.6 & 94.3 & 98.5 & 81.4 & 82.8 \\
\hline
ME & RF-MI & 15 & 81.4 & 98.3 & 99.5 & 83.6 & 85.3 \\
   &    & 30 & 78.4 & 97 & 98.7 & 85.9 & 85.4 \\
   & 2D-CNN-MI & 15 & 69.6 & 90.2 & 97.4 & 84.6 & 80.7 \\
   &    & 30 & 71 & 88.1 & 93.6 & 85 & 80.9 \\
\hline
P8A & RF-MI & 15 & 82 & 97.7 & 100 & 83.9 & 85.7\\
   &    & 30 & 79.3 & 95.9 & 96.8 & 86.5 & 85.8 \\
   & 2D-CNN-MI & 15 & 72.3 & 90.7 & 98.9 & 81.7 & 80.5 \\
   &    & 30 & 74.1 & 91.7 & 97.3 & 83.2 & 81.9 \\
\hline
P8E & RF-MI & 15 & 77.8 & 97.3 & 99.4 & 86.9 & 85.6 \\
   &    & 30 & 81.5 & 96.8 & 99.4 & 85.8 & \ 86.2 \\
   & 2D-CNN-MI & 15 & 74 & 91.9 & 98.3 & 82.9 & 81.6 \\
   &    & 30 & 77 & 92.6 & 98.3 & 83.1 & 82.8 \\
   
\hline
\label{RI levels 1}
\end{tabular}
\end{table}

Table \ref{RI levels 1} presents the performance of the RF and 2D-CNN models under the MI approach across Group~1—the group with the highest number of rainy days as identified in Table \ref{grouped_sites}—and two time windows (15 and 30 minutes), reporting Rec, Spe, and HM metrics. Rec is evaluated across three RI intervals: light rain ($0.1 \leq \mathrm{RI} < 2$ — Rec-L), moderate rain ($2 \leq \mathrm{RI} < 6$ — Rec-M), and heavy rain ($\mathrm{RI} \geq 6$ — Rec-H). Overall, both models exhibit comparable behavior across the two window lengths, with no consistent advantage of one window over the other for the RF model. For instance, while link MT shows a slightly higher HM at 15 minutes (86.1\%) compared to 30 minutes (85.9\%), the remaining links (ME, P8A, and P8E) present very similar or slightly improved HM values when using the 30-minute window. In contrast, the 2D-CNN shows improved performance when using the 30-minute window across links. For instance, at link P8E, the HM improves by about 1.5, with Spe also increasing slightly from 82.9\% to 83.1\%. The Rec values at P8E further confirm this trend, with Rec-L, Rec-M, and Rec-H reaching 77\%, 92.6\%, and 98.3\% for the 30-minute window, compared to 74\%, 91.9\%, and 98.3\% for the 15-minute window. These results indicate that the RF performance remains largely stable across the two window sizes, while the 2D-CNN benefits from the additional temporal context provided by the longer window. Nevertheless, when directly comparing the two models, RF consistently outperforms the 2D-CNN, achieving a better balance between Rec and Spe across all links. For example, on link ME, RF attains an HM of 85.4\%, compared to 80.9\% for the 2D-CNN using the 30-minute window, with higher Rec across all rainfall levels and slightly improved Spe, confirming its overall superiority. The training histories indicate that the 2D-CNN does not suffer from severe overfitting, as confirmed by the small validation–training loss gaps of 0.019, 0.008, 0.034, and 0.020 for ME, MT, P8A, and P8E, respectively, averaged over all days. Moreover, the 2D-CNN achieved higher HM values than the 1D-CNN for all links. For the 30-min window, the 1D-CNN reached HM values of 78.3\%, 79.4\%, 78.4\%, and 77.9\% for ME, MT, P8A, and P8E, respectively, demonstrating the relevance of spatio-temporal information. Although the 2D-CNN benefits from spatio-temporal learning compared with the 1D-CNN, the remaining advantage of the RF model is likely related to the effectiveness of handcrafted features in describing short and irregular rainfall-induced attenuation patterns. Indeed, since rainfall-induced attenuation patterns are irregular and event-dependent rather than strongly periodic, the available dataset may not be sufficient for the CNN to correctly learn sufficiently representative features.

From a computational perspective, RF uses low-dimensional features, leading to low complexity and fast inference. In contrast, the 2D-CNN processes raw data with around 20k parameters and higher cost (845,461 vs. 2,361 FLOPs). Nevertheless, RF performs better, highlighting the effectiveness of engineered features.


\subsection{Multi- and Single Input Approach}

\begin{figure}[b]
\centerline{\includegraphics[width = 1\linewidth]{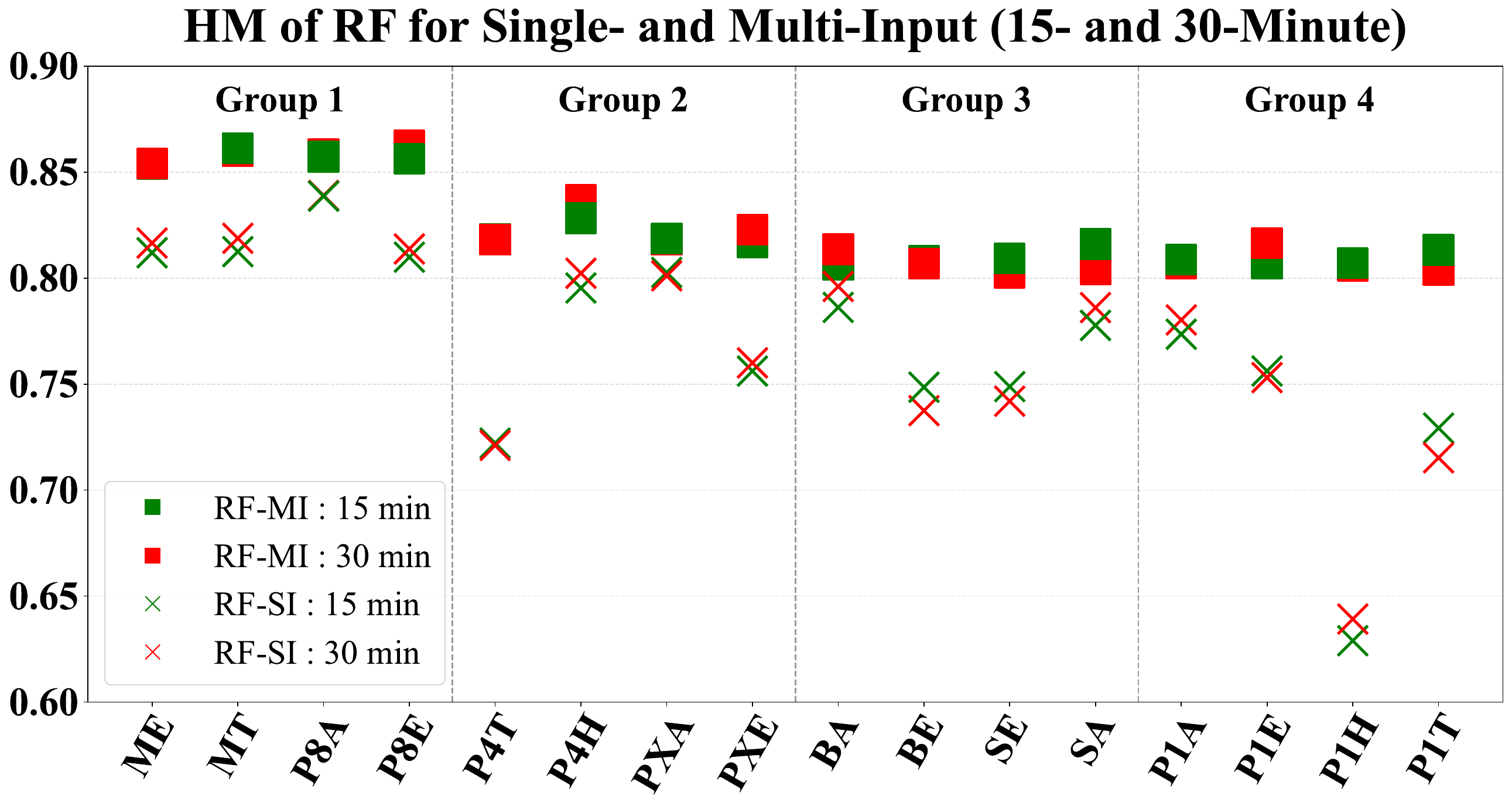}}
\caption{Harmonic mean of RF-based rain detection for single- and multi-input approaches, illustrating performance across 15- and 30-minute time windows. The names on the x axis refer to the single link as listed in Table \ref{grouped_sites}.}
\label{figure-rf}
\end{figure}

Figure \ref{figure-rf} presents an overview of the two approaches, MI and SI, compared for the prediction of wet and dry periods at each link, using the HM as a balanced performance indicator. Both strategies were evaluated using RF with two temporal windows, 15 and 30 minutes, in order to analyze the effect of time-window length on classification performance.

Considering the impact of the time window, RF performance is largely comparable when using either the 15- or 30-minute window across all links for both approaches. the results remain consistent, indicating that window size has a limited influence on overall performance. When comparing the two strategies, MI consistently outperforms SI across all links, with an average improvement of 4–8\%. In addition, the SI approach exhibits less stable behavior, with greater variability in performance across different links. This improvement is observed even for nearby links oriented toward different satellites, indicating that MI can exploit complementary path-integrated observations of the same precipitation field.

Table \ref{RI levels 2} presents the performance comparison between the RF-SI and RF-MI models using 15- and 30-minute windows, reporting Rec across the three RI intervals together with Spe and harmonic mean (HM) for Group~1 (Table \ref{grouped_sites}). For both approaches, the results obtained with the 15- and 30-minute windows remain overall comparable, with only minor variations observed across links and RI levels. Spe remains broadly comparable between the two temporal resolutions, with only minor link-dependent variations. Despite these similarities, RF-MI consistently demonstrates superior overall performance compared to RF-SI. For example, on link ME, RF-MI achieves an HM of 85.3 with a 15-minute window, compared to 81.2 for RF-SI, while on P8E, RF-MI reaches 85.6 versus 80.9 for RF-SI. Consistent improvements are also observed across the Rec metrics. In particular, for link P8A using the 15-minute window, RF-MI outperforms RF-SI across all rainfall intensities, showing a gain of +1.7 for light rainfall and +1.3 for moderate rainfall. Spe further benefits from the multi-input formulation, with RF-MI exceeding 86\% on most links for the 15-minute window, compared to consistently lower values obtained with RF-SI. In addition, quartile-based statistics (Q1, Q2, Q3) were analyzed per group, computed per link and averaged within each group. RF-MI consistently outperforms RF-SI across all groups (Group 1: Q1 77 vs 73.1, Q2 83.4 vs 80.3, Q3 89 vs 85.4; Group 2: Q1 64.6 vs 61.7, Q2 79.4 vs 73.2, Q3 85.9 vs 81.7; Group 3: Q1 63.2 vs 61.1, Q2 79.3 vs 73.2, Q3 88.7 vs 84.6; Group 4: Q1 70.1 vs 58.9, Q2 80.5 vs 71.7, Q3 90.1 vs 80.6), indicating improved typical performance and robustness.

\begin{table}[tb]
\centering
\caption{RF Performance (Rec at Three Rain Levels, Spe, and HM) for Single- and Multi-Input Approaches using 15- and 30-Minute Windows in Group 1.}
\setlength{\tabcolsep}{3pt}
\renewcommand{\arraystretch}{0.9}
\begin{tabular}{c c c c c c c c}
\hline
Link & Model & Window & Rec-L & Rec-M & Rec-H & Spe & HM \\
\hline
MT & RF-MI & 15 & 80 & 98 & 99.5 & 86 & 86.1 \\
   &    & 30 & 79.4 & 97.2 & 99 & 86.4 & 85.9 \\
   & RF-SI & 15 & 76 & 95 & 98.5 & 79.8 & 81.2 \\
   &    & 30 & 76.6 & 94.3 & 97.5 & 80.9 & 81.8 \\
\hline
ME & RF-MI & 15 & 81.4 & 98.3 & 99.5 & 83.6 & 85.3 \\
   &    & 30 & 78.4 & 97 & 98.7 & 85.9 & 85.4 \\
   & RF-SI & 15 & 73.8 & 94.5 & 99.1 & 81.2 & 81.2 \\
   &    & 30 & 73.9 & 94.8 & 96.2 & 82.2 & 81.6 \\
\hline
P8A & RF-MI & 15 & 82 & 97.7 & 100 & 83.9 & 85.7\\
   &    & 30 & 79.3 & 95.9 & 96.8 & 86.5 & 85.8 \\
   & RF-SI & 15 & 80.3 & 96.4 & 100 & 81.7 & 83.8 \\
   &    & 30 & 78.2 & 93.6 & 98.9 & 83.8 & 83.9 \\
\hline

P8E & RF-MI & 15 & 77.8 & 97.3 & 99.4 & 86.9 & 85.6 \\
   &    & 30 & 81.5 & 96.8 & 99.4 & 85.8 & 86.2 \\
   & RF-SI & 15 & 75.7 & 94.5 & 99.4 & 79.7 & 80.9 \\
   &    & 30 & 75.3 & 91.9 & 98.3 & 81.5 & 81.3 \\
   
\hline
\label{RI levels 2}
\end{tabular}
\end{table}

Overall, these results confirm that exploiting spatial correlations improves rainfall discrimination. In particular, RF-MI achieves strong performance for moderate and heavy rainfall while improving light rainfall detection compared to RF-SI. A paired t-test was used to compare MI and SI performance on the same testing days for each link, enabling a direct one-to-one evaluation. Most links exhibit p-values below 0.05, indicating statistically significant improvement with the MI approach. Finally, to assess practical applicability, the MI-RF model was deployed on a Raspberry Pi 4, achieving an inference time of 131 ms per sample, well below the 1-minute sampling interval, confirming real-time capability and enabling edge deployment with reduced communication overhead. Since windows use only past samples, increasing the window size adds no operational lag after the initial warm-up.


\section{Conclusion}
\label{Discussion And Conclusion}

Two ML models, RF and 2D-CNN were evaluated with 15- and 30-min windows under SI and MI configurations, using 2021--2024 microwave-link data from Genoa and radar-derived RI as ground truth. Leave-one-day-out validation was adopted, with performance measured by Rec, Spe, and HM. Results show that MI improves wet/dry detection, while RF provides more stable and reliable performance than 2D-CNN. Despite these promising results, some limitations should be acknowledged. The analysis is restricted to a relatively small geographic area and a limited number of links. Moreover, the findings are based on specific satellite configurations, and their generalization to other regions, link densities, or satellite systems requires further investigation.

Future work will focus on extending the analysis to a wider spatial domain with a larger number of microwave links, as well as investigating how inter-link distance and different link orientations toward different satellites affect classification performance. In addition, strategies to handle missing links, a real-world issue affecting operational robustness, will be explored within the multi-input framework by leveraging available links, an aspect not applicable to the SI approach.

\section{Acknowledgment}
The authors would like to thank the Regional Agency for the
Protection of the Ligurian Environment ARPAL, Italy, for
providing the radar and rain gauges data.

\bibliographystyle{IEEEtran_mod}
\bibliography{bibliography}

\end{document}